\documentclass[11pt]{article}
\usepackage{acl2015}
\usepackage{times}
\usepackage{latexsym}

\usepackage{url}
\usepackage[leqno, fleqn]{amsmath}
\usepackage{amssymb}
\usepackage{qtree}
\usepackage{graphicx}
\usepackage{booktabs}
\usepackage{colortbl}
\usepackage{subcaption}
\usepackage{xcolor}
\usepackage{color}

\newcount\colveccount
\newcommand*\colvec[1]{
        \global\colveccount#1
        \begin{bmatrix}
        \colvecnext
}
\def\colvecnext#1{
        #1
        \global\advance\colveccount-1
        \ifnum\colveccount>0
                \\
                \expandafter\colvecnext
        \else
                \end{bmatrix}
        \fi
}

\newcommand{\nateq}{\equiv}
\newcommand{\natind}{\mathbin{\#}}
\newcommand{\natneg}{\mathbin{^{\wedge}}}
\newcommand{\natfor}{\sqsubset}
\newcommand{\natrev}{\sqsupset}
\newcommand{\natalt}{\mathbin{|}}
\newcommand{\natcov}{\mathbin{\smallsmile}}

\newcommand{\plneg}{\mathop{\textit{not}}}
\newcommand{\pland}{\mathbin{\textit{and}}}
\newcommand{\plor}{\mathbin{\textit{or}}}

\newlength{\howlong}\newcommand{\strikeout}[1]{\settowidth{\howlong}{#1}#1\unitlength0.5ex%
\begin{picture}(0,0)\put(0,1){\line(-1,0){\howlong\divide\unitlength}}\end{picture}}

\newcommand{\True}{\texttt{T}}
\newcommand{\False}{\texttt{F}}
\usepackage{stmaryrd}
\newcommand{\sem}[1]{\ensuremath{\llbracket#1\rrbracket}}

\usepackage{gb4e}
\noautomath

\def\ii#1{\textit{#1}}

\makeatletter
\def\citealt{\def\citename##1{{\frenchspacing##1} }\@internalcitec}
\def\@citexc[#1]#2{\if@filesw\immediate\write\@auxout{\string\citation{#2}}\fi
  \def\@citea{}\@citealt{\@for\@citeb:=#2\do
    {\@citea\def\@citea{;\penalty\@m\ }\@ifundefined
       {b@\@citeb}{{\bf ?}\@warning
       {Citation `\@citeb' on page \thepage \space undefined}}%
{\csname b@\@citeb\endcsname}}}{#1}}
\def\@internalcitec{\@ifnextchar [{\@tempswatrue\@citexc}{\@tempswafalse\@citexc[]}}
\def\@citealt#1#2{{#1\if@tempswa, #2\fi}}
\makeatother

\title{Recursive Neural Networks Can Learn Logical Semantics}

\author{
Samuel R.\ Bowman$^{\ast\dag}$ \\
\texttt{sbowman@stanford.edu} \\[2ex]
$^{\ast}$Stanford Linguistics \\
\And
Christopher Potts$^{\ast}$\\
\texttt{cgpotts@stanford.edu} \\[2ex]
$^{\dag}$Stanford NLP Group
\And
Christopher D.\ Manning$^{\ast\dag\ddag}$\\
\texttt{manning@stanford.edu}\\[2ex]
$^{\ddag}$Stanford Computer Science
}

\date{}

\makeatletter
\newcommand{\@BIBLABEL}{\@emptybiblabel}
\newcommand{\@emptybiblabel}[1]{}
\definecolor{black}{rgb}{0,0,0}
\makeatother
\usepackage[breaklinks, draft, colorlinks, linkcolor=black, urlcolor=black, citecolor=black]{hyperref}

\begin{document}
\maketitle

\begin{abstract} 
  Tree-structured recursive neural networks (TreeRNNs) for sentence meaning have been
  successful for many applications, but it remains an open question whether
  the fixed-length representations that they learn can support tasks as demanding as logical deduction.
  We pursue this question by evaluating whether two such models---plain TreeRNNs and tree-structured neural
  tensor networks (TreeRNTNs)---can correctly learn to identify logical
  relationships such as entailment and contradiction using these representations. 
  In our first set of experiments, we generate artificial data from a
  logical grammar and use it to evaluate the models' ability to learn
  to handle basic relational reasoning, recursive structures, and
  quantification. We then evaluate the models on the more natural
  SICK challenge data. Both models perform competitively on the SICK
  data and generalize well in all three experiments on simulated data,
  suggesting that they can learn suitable representations for logical
  inference in natural language. 
\end{abstract}

\section{Introduction}\label{sec:intro}

Tree-structured recursive neural network models (TreeRNNs; \citealt{goller1996learning}) for sentence meaning
have been successful in an array of sophisticated language tasks,
including sentiment analysis \cite{socher2011semi,irsoydeep},
image description \cite{sochergrounded}, and paraphrase detection
\cite{Socher-etal:2011:Paraphrase}. These results are encouraging for
the ability of these models to learn to produce and use strong 
semantic representations for sentences. However, it remains an 
open question whether any such fully learned model can achieve the kind of
high-fidelity distributed representations proposed in recent algebraic
work on vector space modeling 
\cite{ClarkCoeckeSadrzadeh2011,grefenstette2013towards,Hermann-etal:2013,rocktaschellow},
and whether any such model can match the performance of grammars based in logical forms
in their ability to model core semantic phenomena like quantification, entailment, and contradiction \cite{Warren:Pereira:1982,Zelle:Mooney:1996,ZetCol:2005,LiangJordan:2013}.

Recent work on the algebraic approach of
\newcite{ClarkCoeckeSadrzadeh2011} has yielded rich frameworks 
for computing the meanings of fragments of natural
language compositionally from vector or tensor representations, but 
has not yet yielded effective methods for learning these representations
from data in typical machine learning settings. 
Past experimental work on reasoning with
distributed representations have been largely confined to short
phrases \cite{Mitchell:Lapata:2010,Grefenstette-etal:2011,baroni2012entailment}.
However, for robust natural language understanding, it is essential to
model these phenomena in their full generality on complex linguistic
structures. 

This paper describes four machine learning experiments
that directly evaluate the abilities of these models to learn representations that 
support specific semantic behaviors. These tasks follow the format of
 \ii{natural language inference} 
(also known as \ii{recognizing textual entailment};
\citealt{dagan2006pascal}), in which the goal is to determine the core
inferential relationship between two sentences. 
We introduce a novel NN architecture for natural language inference which
independently computes vector representations for each of two sentences using
standard TreeRNN or TreeRNTN \cite{socher2013acl1} models, and
produces a judgment for the pair using only those representations. This allows
us to gauge the abilities of these two models to represent all of the necessary semantic
information in the sentence vectors. 

Much of the
theoretical work on natural language inference (and some successful implemented models;
\citealt{maccartney2009extended,watanabe2012latent}) involves \ii{natural
  logics}, which are formal systems that define rules of inference
between natural language words, phrases, and sentences without the
need of intermediate representations in an artificial logical
language.
In our first three experiments, we test our models' ability to learn the foundations of natural
language inference by training them to reproduce the behavior of the natural logic of \newcite{maccartney2009extended} 
on artificial data. This logic
defines seven mutually-exclusive relations of synonymy, entailment, contradiction,
and mutual consistency, as summarized in Table~\ref{b-table}, and it
provides rules of semantic combination for projecting these relations
from the lexicon up to complex phrases. The formal properties of this system 
are now well-understood \cite{Icard:Moss:2013,Icard:Moss:2013:LILT}.
The first experiment using this logic 
covers reasoning with the bare logical relations (\S\ref{sec:join}), the second extends this to reasoning with statements constructed compositionally from recursive functions (\S\ref{sec:recursion}),
and the third covers the additional complexity that results from quantification (\S\ref{sec:quantifiers}).
Though the performance of the plain TreeRNN model 
 is somewhat poor in our first experiment, we find that the stronger TreeRNTN model
  generalizes well in every case, suggesting that it has learned to simulate our target logical concepts.

The experiments with simulated data provide a convincing demonstration
of the ability of neural networks to learn to build and use semantic representations for complex natural language sentences from reasonably-sized
training sets. However, we are also interested in the more practical
question of whether they can learn these representations from
naturalistic text. To address this question, we apply our models to
the SICK entailment challenge data in \S\ref{sec:sick}. The small size of this corpus puts
data-hungry NN models like ours at a disadvantage, but we are
nonetheless able to achieve competitive performance on it,
surpassing several submitted models with significant hand-engineered task-specific features and our own NN baseline.
This suggests that the representational abilities that we observe in the previous sections
are not limited to carefully circumscribed tasks. We conclude that TreeRNTN models
are adequate for typical cases of natural language inference, and that there is not yet
any clear level of inferential complexity for which other approaches work and NN models fail.

\begin{table*}[tp]
  \centering\small
  \setlength{\tabcolsep}{15pt}
  \renewcommand{\arraystretch}{1}
  \begin{tabular}{l c l l} 
    \toprule
    Name & Symbol & Set-theoretic definition & Example \\ 
    \midrule
    (strict) entailment         & $x \natfor y$   & $x \subset y$ & \ii{turtle, reptile}  \\ 
    (strict) reverse entailment & $x \natrev y$   & $x \supset y$ & \ii{reptile, turtle}  \\ 
    equivalence        & $x \nateq y$    & $x = y$       & \ii{couch, sofa} \\ 
    alternation        & $x \natalt y$   & $x \cap y = \emptyset \wedge x \cup y \neq \mathcal{D}$ & \ii{turtle, warthog} \\ 
    negation           & $x \natneg y$   & $x \cap y = \emptyset \wedge x \cup y = \mathcal{D}$    & \ii{able, unable} \\
    cover              & $x \natcov y$   & $x \cap y \neq \emptyset \wedge x \cup y = \mathcal{D}$ & \ii{animal, non-turtle} \\ 
    independence       & $x \natind y$   & (else) & \ii{turtle, pet}\\
    \bottomrule
  \end{tabular}
  \caption{\label{b-table}The seven relations of MacCartney and Manning \protect\shortcite{maccartney2009extended}'s logic are defined abstractly on pairs of sets drawing from the universe $\mathcal{D}$, but can be straightforwardly applied to any pair of natural language words, phrases, or sentences. The relations are defined so as to be mutually exclusive.} 
 
\end{table*}

\section{Tree-structured neural networks} \label{methods}

We limit the scope of our experiments in this paper to neural network models 
that adhere to the linguistic \ii{principle of
 compositionality}, which says that the meanings for complex
expressions are derived from the meanings of their parts
via specific composition functions \cite{Partee84,Janssen97}. In our
distributed setting, word meanings are embedding vectors of dimension $n$. A learned
composition function maps pairs of them to single phrase vectors of dimension $n$, 
which can then be merged again to represent more complex
phrases, forming a tree structure. Once the entire sentence-level representation has been
derived at the top of the tree, it serves as a fixed-dimensional input for some subsequent layer function.

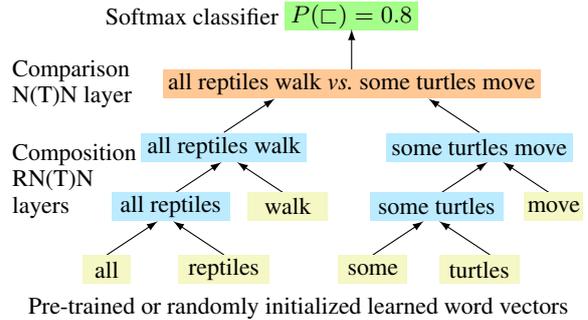
\begin{figure}[tp]
  \centering
  \footnotesize

\newcommand{\labeledtreenode}[4][3.5]{\put(#2){\makebox(0,0){{\colorbox{#4}{\makebox(#1,0.3){#3}}}}}}

\newcommand{\textlabel}[4][3.5]{\put(#2){\makebox(0,0){\makebox(#1,0.3){#3}}}}

\definecolor{lexcolor}{HTML}{F5F7C4}
\definecolor{compositioncolor}{HTML}{BBEBFF}
\definecolor{comparisoncolor}{HTML}{FFC895}
\definecolor{softmaxcolor}{HTML}{A5FF8A}

  

      

  
    




  




          
  

\setlength{\unitlength}{0.56cm}
\begin{picture}(13,7.5)
  
  \labeledtreenode[2.75]{7.75,7}{$P(\sqsubset) = 0.8$}{softmaxcolor}  
  \put(7.75,5.7){\vector(0,1){1}}  
  \labeledtreenode[8.5]{7.75,5.4}{all reptiles walk \emph{vs.}~some turtles move}{comparisoncolor}

  \textlabel{4,7}{Softmax classifier}{black}
  \textlabel{1.25,5.4}{\parbox{0.65in}{Comparison N(T)N layer}}{black}      
  \textlabel{1.25,3.1}{\parbox{0.65in}{Composition RN(T)N layers}}{black}
  \textlabel{6.5,0.1}{Pre-trained or randomly initialized learned word vectors}{black}
  
    
  \put(2,1.35){\vector(3,2){1.2}}
  \labeledtreenode[0.75]{2,1}{all}{lexcolor}

  \put(4.75,1.35){\vector(-3,2){1.2}}
  \labeledtreenode[1.75]{4.75,1}{reptiles}{lexcolor}

  \put(3.5,2.75){\vector(3,2){1.2}}
  \labeledtreenode[2.4]{3.5,2.5}{all reptiles}{compositioncolor}

  \put(6.25,2.75){\vector(-3,2){1.2}}
  \labeledtreenode[1.5]{6.25,2.5}{walk}{lexcolor}

  \put(4.75,4.25){\vector(3,2){1.2}}
  \labeledtreenode{4.75,3.9}{all reptiles walk}{compositioncolor}
  

  \put(8.25,1.35){\vector(3,2){1.2}}
  \labeledtreenode[1.25]{8.25,1}{some}{lexcolor}

  \put(11,1.35){\vector(-3,2){1.2}}
  \labeledtreenode[1.5]{10.75,1}{turtles}{lexcolor}

  \put(9.75,2.75){\vector(3,2){1.2}}
  \labeledtreenode[2.75]{9.75,2.5}{some turtles}{compositioncolor}

  \put(12.5,2.75){\vector(-3,2){1.2}}
  \labeledtreenode[1]{12.5,2.5}{move}{lexcolor}
          
  \put(10.75,4.25){\vector(-3,2){1.2}}
  \labeledtreenode[4]{10.75,3.9}{some turtles move}{compositioncolor}
  
\end{picture}
  \caption{In our model, two separate tree-structured networks build up vector representations for each of two sentences using either NN or NTN layer functions. A comparison layer then uses the resulting vectors to produce features for a classifier.} 
  \label{sample-figure}
\end{figure}

To apply these recursive models to our task, we propose the tree 
pair model architecture depicted in Fig.~\ref{sample-figure}. 
In it, the two phrases being compared are processed separately using a pair 
of tree-structured networks that share a single set of parameters. 
The resulting vectors are fed into a separate comparison
layer that is meant to generate a feature vector capturing the
relation between the two phrases. The output of this layer is then
given to a softmax classifier, which produces a
distribution over the seven relations represented in Table~\ref{b-table}.

For the sentence embedding portions of the network, we evaluate both TreeRNN models with
 the standard NN layer function \eqref{TreeRNN} and those with the more powerful neural tensor 
 network layer function
\eqref{TreeRNTN} proposed in \newcite{chen2013learning}. The nonlinearity $f(x) = \tanh(x)$ is applied
 elementwise to the output of either layer function.
\begin{gather} 
\label{TreeRNN}
\vec{y}_{\textit{TreeRNN}} = f(\mathbf{M} \colvec{2}{\vec{x}^{(l)}}{\vec{x}^{(r)}} + \vec{b}\,) \\
\label{TreeRNTN} 
\vec{y}_{\textit{TreeRNTN}} = \vec{y}_{\textit{TreeRNN}} + f(\vec{x}^{(l)T} \mathbf{T}^{[1 \ldots n]} \vec{x}^{(r)})
\end{gather} 
Here, $\vec{x}^{(l)}$ and $\vec{x}^{(r)}$ are the column vector
representations for the left and right children of the node, and
$\vec{y}$ is the node's output.  The TreeRNN concatenates them, multiplies
them by an $n \times 2n$ matrix of learned weights, and adds a bias $\vec{b}$. 
The TreeRNTN adds a learned full rank third-order tensor 
$\mathbf{T}$, of dimension $n \times n \times n$, modeling
multiplicative interactions between the child vectors. 
The comparison layer uses the same layer function as the
composition layers (either an NN layer or an NTN layer) with
independently learned parameters and a separate nonlinearity function.
Rather than use a $\tanh$ nonlinearity here, we found better results with the leaky rectified linear function
\cite{maasrectifier}: $f(x)=\max(x, 0) +
0.01\min(x, 0)$. 

Other strong tree-structured models have been proposed in past work
\cite{sochergrounded,irsoydeep,tai2015improved}, but
we believe that these two provide a valuable case study, and that positive results on 
here are likely to generalize well to stronger models.

To run the model forward, we assemble the two tree-structured networks so as to match the structures provided for each phrase, which are either included in the source data or given by a parser.
The word vectors are then looked up from the vocabulary embedding matrix $V$ (one of the learned model parameters), and
the composition and comparison functions are used to pass information
up the tree and into the classifier. For an objective
function, we use the negative log likelihood of the
correct label with tuned L2 regularization.

We initialize parameters uniformly, using the range $(-0.05, 0.05)$ for layer parameters
and $(-0.01, 0.01)$ for embeddings, and train the model using stochastic gradient descent (SGD)
with learning rates computed using AdaDelta \cite{zeiler2012adadelta}.
The classifier feature vector is fixed at 75 dimensions and 
the dimensions of the recursive layers are tuned manually.
Training times on CPUs vary from hours to days across experiments.
On the experiments which use artificial data, we report mean
results over five fold cross-validation, where
variance across runs is typically no
more than two percentage points. In addition, because the classes are not necessarily balanced, 
we report both accuracy and macroaveraged F1.\footnote{We compute macroaveraged F1 
as the harmonic mean of average precision and average recall, both computed
for all classes for which there is test data, setting precision to 0 
where it is not defined.}
Source code and generated data will be released after the review period.

\begin{table}[tp]
  \centering  \small
  \setlength{\arraycolsep}{8pt}
  \renewcommand{\arraystretch}{1.1}
  \newcommand{\UNK}{\cdot}  
  $\begin{array}[t]{c@{ \ }|*{7}{c}|}
    \multicolumn{1}{c}{}
             & \nateq     & \natfor     & \natrev     & \natneg    & \natalt     & \natcov     & \multicolumn{1}{c}{\natind} \\
    \cline{2-8}
    \nateq  & \nateq &   \natfor &  \natrev &  \natneg &   \natalt &  \natcov &  \natind \\
    \natfor & \natfor &  \natfor &  \UNK &  \natalt &   \natalt &  \UNK &  \UNK \\
    \natrev & \natrev &  \UNK &  \natrev &  \natcov &   \UNK &  \natcov &  \UNK \\
    \natneg & \natneg &  \natcov &  \natalt &  \nateq &    \natrev &  \natfor &  \natind \\
    \natalt & \natalt &  \UNK &  \natalt &  \natfor &   \UNK &  \natfor &  \UNK \\
    \natcov & \natcov &  \natcov &  \UNK &  \natrev &   \natrev &  \UNK &  \UNK \\
    \natind & \natind & \UNK &  \UNK &  \natind &  \UNK &  \UNK &  \UNK \\
    \cline{2-8}
  \end{array}$
  \caption{In \S\ref{sec:join}, we assess our models' ability to learn to do inference over pairs of relations using the rules represented here, which are derived from the definitions of the relations in Table~\ref{b-table}.  As an example, given that $p_1 \natfor p_2$ and $p_2 \natneg p_3$, the entry in the $\natfor$ row and the $\natneg$ column lets us conclude that $p_1 \natalt p_3$. Cells containing a dot correspond to situations for which no valid inference can be drawn.} 
  \label{tab:jointable}
\end{table}

\section{Reasoning about semantic relations}\label{sec:join}

The simplest kinds of deduction in natural logic involve atomic statements 
using the relations in Table~\ref{b-table}. 
For instance, from the relation $p_1 \natrev p_2$ between two propositions, 
one can infer the relation $p_2 \natfor p_1$ by applying the definitions of the relations directly. 
If one is also given the relation $p_2 \natrev p_3$ one can conclude that $p_1 \natrev p_3$, by basic set-theoretic reasoning (transitivity of $\natrev$). The
full set of sound such inferences on pairs of premise relations is depicted in
Table~\ref{tab:jointable}. Though these basic inferences do not involve compositional
sentence representations, any successful reasoning using compositional representations
will rely on the ability to perform sound inferences of this kind, so our first experiment studies how well each model can learn to perform them them in isolation.


\paragraph{Experiments}
We begin by creating a world model
on which we will base the statements in the train and test sets.
This takes the form of a small Boolean structure in which terms denote
sets of entities from a small domain.  Fig.~\ref{lattice-figure}a
depicts a structure of this form with three entities ($a$, $b$, and $c$) and eight proposition terms ($p_1$--$p_8$). We then generate a 
relational statement for each pair of terms in the model, as shown in Fig.~\ref{lattice-figure}b. 
We divide these statements evenly into train and test sets, and delete the test set
 examples which cannot be proven from the train examples, for which there is not enough information for even an ideal system to choose a correct label.
In each experimental run, we create a model with 80 terms over a domain of 7 elements, yielding a training set of 3200 examples and a test set of 
2960 examples.

We trained models with both the NN and NTN comparison functions on these
data sets.\footnote{Since this task relies crucially on the learning of a pair of vectors, no simpler version of our model is a viable baseline.} 
In both cases, the models are implemented as
described in \S\ref{methods}, but since the items being compared
are single terms rather than full tree structures, the composition
layer is not used, and the two models are not recursive. We simply present
the models with the (randomly initialized) embedding vectors for each
of two terms, ensuring that the model has no information about the terms
being compared except for the relations between them that appear in training.

\begin{figure}[t]
  \centering
  \begin{subfigure}[t]{0.45\textwidth}
    \centering
    \newcommand{\labelednode}[4]{\put(#1,#2){\oval(1.5,1)}\put(#1,#2){\makebox(0,0){$\begin{array}{c}#3\\\{#4\}\end{array}$}}}
    \setlength{\unitlength}{1cm}\scalebox{0.8}{
    \begin{picture}(5,5.5)
      \labelednode{2.50}{5}{}{a,b,c}
      
      \put(0.75,4){\line(3,1){1.5}}
      \put(2.5,4){\line(0,1){0.5}}
      \put(4.25,4){\line(-3,1){1.5}}
      
      \labelednode{0.75}{3.5}{p_1,p_2}{a,b}
      \labelednode{2.50}{3.5}{p_3}{a,c}
      \labelednode{4.25}{3.5}{p_4}{b,c}
      
      \put(0.75,2.5){\line(0,1){0.5}}
      \put(0.75,2.5){\line(3,1){1.5}}
      
      \put(2.5,2.5){\line(-3,1){1.5}}
      \put(2.5,2.5){\line(3,1){1.5}}
      
      \put(4.25,2.5){\line(0,1){0.5}}
      \put(4.25,2.5){\line(-3,1){1.5}}

      \labelednode{0.75}{2}{p_5,p_6}{a}
      \labelednode{2.50}{2}{}{b}
      \labelednode{4.25}{2}{p_7,p_8}{c}
      
      \put(2.5,1){\line(-3,1){1.5}}
      \put(2.5,1){\line(0,1){0.5}}
      \put(2.5,1){\line(3,1){1.5}}
      
      \labelednode{2.5}{0.5}{}{}
    \end{picture}}
    \caption{Example boolean structure. The terms $p_1$--$p_8$ name the sets. Not all sets have names, and  some sets have multiple names, so that learning $\nateq$ is non-trivial.}
  \end{subfigure}
  \qquad\small
    \begin{subfigure}[t]{0.43\textwidth}
    \centering \vspace{0.4cm}
    \setlength{\tabcolsep}{12pt}
    \begin{tabular}[b]{c  c}
      \toprule
      Train & Test \\
      \midrule
      $p_1 \nateq p_2$              & $p_2 \natneg p_7$ \\
      $p_1 \natrev p_5$              & $p_2 \natrev p_5$ \\
      $p_4 \natrev p_8$              & \strikeout{$p_5 \nateq p_6$} \\
      $p_5 \natalt p_7$              & \strikeout{$p_7 \natfor p_4$} \\
      $p_7 \natneg p_1$           & $p_8 \natfor p_4$ \\

      \bottomrule
    \end{tabular}

    \caption{A few examples of atomic statements about the
      model.  Test statements that are not provable from the training data shown are
      crossed out.}
  \end{subfigure}  
  \caption{Small example structure and data for learning relation composition.}
  \label{lattice-figure}
\end{figure} 

\begin{table}[tp]
  \centering\small
  \begin{tabular}{ l r@{ \ }r r@{ \ }r }
    \toprule
    ~&\multicolumn{2}{c}{Train} & \multicolumn{2}{c}{Test}\\
    \midrule
    $\natind$ only &53.8 &(10.5)    &53.8 &(10.5) \\
    15d NN &				99.8&	(99.0) &94.0&(87.0) \\
    15d NTN 				& \textbf{100} & \textbf{(100)} & \textbf{99.6} & \textbf{(95.5)}\\
    \bottomrule
  \end{tabular}

  \caption{Performance on the semantic relation experiments. These results and all other results on artificial data are reported as mean accuracy scores over five runs followed by mean macroaveraged F1 scores in parentheses. The ``$\natind$ only'' entries reflect the frequency of the most frequent class.}
  \label{joinresultstable}
\end{table}

\paragraph{Results} 
The resuts (Table \ref{joinresultstable}) show that NTN is able to accurately encode the relations between the terms in the geometric relations between their vectors, 
and is able to then use that information to recover relations that 
are not overtly included in the training data. The NN also generalizes fairly well, 
but makes enough errors that it remains an open question whether 
it is capable of learning representations with these properties. 
It is not possible for us to rule out the possibility that different optimization techniques or
further hyperparameter tuning could lead an NN model to succeed here.

As an example from our test data, both models correctly labeled $p_1 \natfor p_3$, potentially learning from the training examples $\{p_1 \natfor p_{51},~p_3 \natrev p_{51}\}$ or $\{p_1\natfor p_{65},~p_3 \natrev p_{65} \}$. On another example involving comparably frequent relations, the NTN correctly labeled $p_6 \natrev p_{24}$, likely on the basis of the training examples $\{p_6 \natcov p_{28},~p_{28} \natneg p_{24}\}$, while the NN incorrectly assigned it $\natind$.

\section{Recursive structure}\label{sec:recursion}

A successful natural language inference system must reason 
about relations not just over familiar
atomic symbols, but also over novel structures built up 
recursively from these symbols. 
This section shows that our models can learn a 
compositional semantics over such structures.
In our evaluations, we exploit the fact that our logical language
is infinite by testing on strings that are longer and more complex
than any seen in training.


\paragraph{Experiments}
As in \S\ref{sec:join}, we generate artificial data from a formal system,
 but we now replace the unanalyzed symbols
from that experiment with complex formulae. These formulae
represent a complete classical propositional logic:
each atomic symbol is a variable over the domain \{$\True$, $\False$\}, and the only
operators are truth-functional ones.  Table~\ref{tab:pl} defines this
logic, and Table~\ref{tab:plexs} gives some short examples of relational statements from our data.
 To compute these relations
between statements, we exhaustively enumerate the sets of assignments
of truth values to propositional variables that would satisfy each of
the statements, and then we convert the set-theoretic relation between
those assignments into one of the seven relations in
Table~\ref{b-table}. As a result, each relational statement represents
a valid theorem of the propositional logic, and to succeed, the models 
must learn to reproduce the behavior of a theorem prover.\footnote{
Socher et al.~\shortcite{socher2012semantic} show that a matrix-vector TreeRNN
model somewhat similar to our TreeRNTN can learn boolean logic, 
a logic where the atomic symbols are simply the
values $\True$ and $\False$. While learning the operators of that logic is not trivial, the outputs of
each operator can be represented accurately by a single bit.
In the much more demanding task presented here, the atomic symbols are variables over these values, and the sentence vectors must thus be able to distinguish up to $2^{64}$ distinct conditions on valuations.}

\begin{table}[tp]
  \centering\small
  \begin{subtable}[t]{0.45\textwidth}
    \centering
    \begin{tabular}[t]{l l}
      \toprule
      Formula     & Interpretation \\
      \midrule
      $p_1$, $p_2$, $p_3$, $p_4$, $p_5$, $p_6$ & $\sem{x} \in \{\True, \False\}$ \\
      $\plneg \varphi$ & $\True$ iff $\sem{\varphi} = \False$ \\
      $(\varphi \pland \psi)$ & $\True$ iff $\False \notin \{\sem{\varphi}, \sem{\psi}\}$ \\
      $(\varphi \plor \psi)$  & $\True$ iff $\True \in \{\sem{\varphi}, \sem{\psi}\}$ \\
      \bottomrule
    \end{tabular}    
    \caption{Well-formed formulae. $\varphi$ and $\psi$
      range over all well-formed formulae, and $\sem{\cdot}$ is
      the interpretation function mapping formulae into $\{\True,
      \False\}$.}\label{tab:pl}
  \end{subtable}
  \begin{subtable}[t]{0.45\textwidth}
    \centering\vspace{4mm}
    \begin{tabular}[t]{r c l}
      \toprule
      $\plneg p_3$        & $\natneg$ & $p_3$ \\
      $\plneg \plneg p_6$ & $\nateq$  & $p_6$ \\
      $p_3$               & $\natfor$ & $(p_3 \plor p_2)$ \\
      $(p_1 \plor (p_2 \plor p_4))$               & $\natrev$ & $(p_2 \pland  \plneg p_4)$ \\
      $\plneg\, (\plneg p_1 \pland \plneg p_2)$ & $\nateq$ & $(p_1 \plor p_2)$ \\ 
      \bottomrule
    \end{tabular}
    \caption{Examples of the type of statements used for training and testing. These are relations between
      well-formed formulae, computed in terms of sets of satisfying
      interpretation functions $\sem{\cdot}$.}\label{tab:plexs}
  \end{subtable}
  \caption{Natural logic relations over sentences of propositional logic.}  
  \label{prop-figure}
\end{table}

In our experiments, we randomly generate unique pairs 
of formulae containing up to 12 instances of logical operators each and compute the relation that holds for each pair.
We discard pairs in which either statement is either a tautology or a
contradiction, for which the seven relations in
Table~\ref{b-table} are undefined. The resulting set of formula pairs is
then partitioned into 12 bins according the number of operators in
the larger of the two formulae. We then sample 20\% of each
bin for a held-out test set.
If we do not implement any constraint that the two statements being
compared are similar in any way, then the generated data are dominated
by statements in which the two formulae refer to largely separate
subsets of the six variables, which means that the $\natind$ relation
is almost always correct.  In an effort to balance the distribution of
relation labels without departing from the basic task of modeling
propositional logic, we disallow individual pairs of statements from
referring to more than four of the six propositional variables.

In order to test the model's generalization to unseen structures, we discard
training examples with more than 4 logical operators, yielding 60k short training examples,
and 21k test examples across all 12 bins.
In addition to the two tree models, we also train a summing NN baseline
which is largely identical to the TreeRNN, except that instead of using a learned composition function,
it simply sums the term vectors in each expression to compose them before passing them to the comparison layer. Unlike the two tree models, this baseline does not use word order,
and is as such guaranteed to ignore some information that it would need in order to succeed perfectly.

\begin{figure}[t]
  \centering
  \includegraphics[width=3.05in]{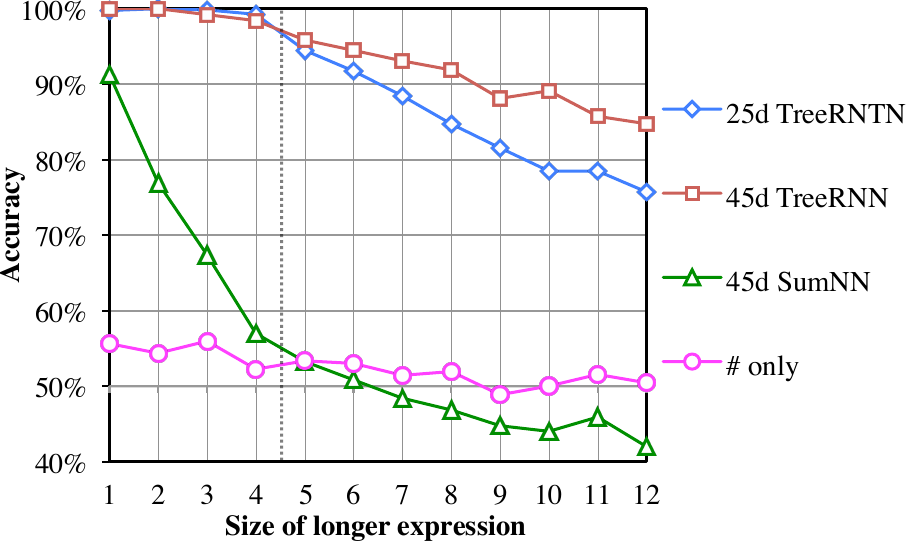}
  \caption{Results on recursive structure. The vertical dotted line marks the size of the longest training examples.}
    
  \label{prop-results} 
\end{figure}

\paragraph{Results} Fig.~\ref{prop-results} shows the relationship
between test accuracy and statement size. While the summing baseline model performed poorly across the board, we found that both recursive
models were able to perform well on unseen small test examples, 
with TreeRNN accuracy above
98\% and TreeRNTN accuracy above 99\% on formulae below length five, indicating
that they learned correct approximations of the underlying
logic. Training accuracy was 66.6\% for the SumNN, 99.4\% for the TreeRNN, and 99.8\% for the TreeRNTN.

After the size four training cutoff, performance gradually decays with expression size for both tree models, suggesting that the learned approximations were accurate but lossy.
Despite the TreeRNTN's stronger performance on short sentences, its performance
decayed more quickly than the TreeRNN's. 
This suggests to us that it learned to interpret many specific fixed-size tree structures directly,
allowing it to get away without learning as robust generalizations about how to compose
terms in the general case.
Two factors may have contributed to the learning of these narrower generalizations: 
even with the lower dimension,
the TreeRNTN composition function has about eight times as many parameters as the
TreeRNN, and the TreeRNTN worked best with weaker L2 regularization than the 
TreeRNN ($\lambda = 0.0003$ vs. $0.001$). 
However, even in the most complex set of test examples, the TreeRNTN classifies true examples of every
 class but $\nateq$ (which is rare in long examples, and occurs only once here) correctly 
the majority of the time, and
 the performance of both models on those examples indicates that both have learned
  reasonable approximations of the underlying theorem proving task over recursive structure.

\section{Reasoning with quantifiers and negation}\label{sec:quantifiers}

We have seen that recursive models can learn an approximation of propositional
logic.  However, natural languages can express functional meanings of
considerably greater complexity than this.  As a key test of whether our models 
can capture this complexity, we now study the degree to which they are able to
develop suitable representations for the semantics of natural language
quantifiers like \ii{most} and \ii{all} as they interact with negation and lexical entailments. Quantification 
and negation are far from the only place in natural language where complex functional meanings
are found, but they are natural focus, since they have
formed a standard case study in prior formal work on natural
language inference \cite{Icard:Moss:2013:LILT}.

\paragraph{Experiments}
Our data consist of pairs of sentences generated
from a grammar for a simple English-like artificial language.
Each sentence contains a quantifier, a noun
which may be negated, and an intransitive verb which may be
negated. We use the quantifiers \ii{some}, \ii{most}, \ii{all},
\ii{two}, and \ii{three}, and their negations \ii{no}, \ii{not-all},
\ii{not-most}, \ii{less-than-two}, and \ii{less-than-three}, and also
include five nouns, four intransitive verbs, and the negation symbol
\ii{not}. In order to be able to define relations between sentences
with differing lexical items, we define the lexical relations for
each noun--noun pair, each verb--verb pair, and each
quantifier--quantifier pair. The grammar then generates pairs of
sentences and calculates the relations
between them.  
For instance, our models might then see
pairs like \eqref{p1} and \eqref{p2} in training and be required to 
then label \eqref{p3}.

\vspace{-0.6cm}
\begin{gather}
  \text{(most turtle) swim} \natalt \text{(no turtle) move}\label{p1}
  \\
  \text{(all lizard) reptile} \natfor  \text{(some lizard) animal}\label{p2}
  \\
  \text{(most turtle) reptile} \natalt \text{(all turtle) (not animal)}\label{p3}
\end{gather}



In each run, we randomly partition the set of valid \textit{single sentences} under the grammar into training and test, and then label all of the pairs from within each set to generate a training set of 27k pairs and a test set of 7k pairs. Because the model doesn't see the test sentences at training time, it cannot directly use the kind of reasoning described in \S\ref{sec:join}, and must instead both infer the word-level relations and learn a complete reasoning system over them for our logic. 

We use the same summing baseline as in \S\ref{sec:recursion}.
The highly consistent  sentence structure in this experiment means that this model
is not as disadvantaged by the lack of word order information as it is in the previous experiment, 
but the variable placement of \ii{not}  nonetheless introduces potential uncertainty in the 58.8\% 
of examples that contain a sentence with a single token of it.

\begin{table}[tp]
  \centering\small
    \begin{tabular}{ l r@{ \ }r r@{ \ }r }
    
    \toprule
    ~ & \multicolumn{2}{c}{Train} & \multicolumn{2}{c}{Test} \\
    \midrule
    $\natind$ only &	35.4 & (7.5)	& 35.4	& (7.5)\\
    25d SumNN	&	96.9&	(97.7)&	93.9&	(95.0)\\	
    25d TreeRNN	&	99.6&	(99.6)&	99.2&	(99.3)\\
    25d TreeRNTN	&	\textbf{100}&(\textbf{100})&	\textbf{99.7} & \textbf{(99.5)}\\
    \bottomrule
  \end{tabular}
  
  \caption{Performance on the quantifier experiments, given as \% correct and macroaveraged F1.}
  \label{qresultstable}
\end{table} 

%
%
\paragraph{Results} The results (Table~\ref{qresultstable}) show that both tree models are able to learn to generalize the underlying logic almost perfectly. The baseline summing model can largely memorize the training data, but does not generalize as well. We do not find any consistent pattern in the handful of errors made by either tree model, and no errors were consistent across model restarts, suggesting that there is no fundamental obstacle to learning a perfect model for this problem.

\section{The SICK textual entailment challenge}\label{sec:sick}


The specific model architecture that we use is novel, and though the underlying tree structure approach has been validated elsewhere, our experiments so far do not guarantee that it viable model for handling inference over real
natural language data. To investigate our models' ability to handle the noisy labels and the diverse range of linguistic structures seen in typical natural language data, we use the SICK textual entailment challenge corpus \cite{marelli2014sick}. The corpus consists of about 10k natural language sentence pairs, labeled with \ii{entailment}, \ii{contradiction}, or \ii{neutral}. At only a few thousand distinct sentences (many of them variants on an even smaller set of template sentences), the corpus is not large enough to train a high quality learned model of general natural language, but it is the largest human-labeled entailment corpus that we are aware of, and our results nonetheless show that tree-structured NN models can learn to do inference in the real world.

Adapting to this task requires us to make a few additions to the techniques discussed in \S\ref{methods}. In order to better handle rare words, we initialized our word embeddings using 200 dimensional vectors trained with 
GloVe \cite{pennington2014glove} on data from Wikipedia. Since 200 dimensional vectors are too large to be practical in an TreeRNTN on a small dataset, a new embedding transformation layer is needed. Before any embedding is used as an input to a recursive layer, it is passed through an additional $\tanh$ neural network layer with the same output dimension as the recursive layer. This new layer aggregates any usable information from the embedding vectors into a more compact working representation. An identical layer is added to the SumNN between the word vectors and the comparison layer.

We also supplemented the SICK training data\footnote{We tuned the model using performance on a held out development set, but report performance here for a version of the model trained on both the training and development data and tested on the 4,928 example SICK test set. We also report training accuracy on a small sample from each data source.} with 600k examples of entailment data from the Denotation Graph project (DG, \citealt{hodoshimage}, also used by the winning SICK submission), a corpus of noisy automatically labeled entailment examples over image captions, the same genre of text from which SICK was drawn. We trained a single model on data from both sources, but used a separate set of softmax parameters for classifying into the labels from each source. We parsed the data from both sources with the Stanford PCFG Parser v.~3.3.1 \cite{klein2003accurate}. We also found that we were able to train a working model much more quickly with an additional technique: we collapse subtrees that were identical across both sentences in a pair by replacing them with a single head word. The training and test data on which we report performance are collapsed in this way, and both collapsed and uncollapsed copies of the training data are used in training. Finally, in order to improve regularization on the noisier data, we used dropout \cite{srivastava2014dropout} at the input to the comparison layer (10\%) and at the output from the embedding transform layer (25\%). 

\begin{table}[tp]
  \centering \small
    \begin{tabular}{ l@{\hspace{-0.25em}} r@{~~~~} r@{~~~~} r@{~~~~} r@{~~~~} }
    \toprule
        ~&\ii{neutral}&	 30d  & 			30d & 50d\\
    ~&only &SumNN  &TrRNN &TrRNTN\\ 
     \midrule
    DG Train	& 50.0 & 68.0 & 67.0 & \textbf{74.0} \\
    SICK Train	& 56.7 & 96.6 & 95.4 & \textbf{97.8} \\
    SICK Test	& 56.7 & 73.4 & 74.9 & \textbf{76.9} \\
    \midrule
    \textsc{Passive} (4\%)	& 0 		& 76  		& 68		&\textbf{88}\\   
    \textsc{Neg} (7\%)		& 0 		& 96	 		& \textbf{100} & \textbf{100}\\
    \textsc{Subst} (24\%)	& 28 		& \textbf{72}  		& 64 		&  \textbf{72}\\
    \textsc{MultiEd} (39\%)	&  \textbf{68} & 61  		&66 		& 64 \\
    \textsc{Diff} (26\%)		& \textbf{96} &  	68		&79		& \textbf{96}\\  
    \midrule
    \textsc{Short} (47\%) & 50.0 & 73.9 & 73.5		& \textbf{77.3} \\    
    \bottomrule
  \end{tabular}
  \caption{Classification accuracy, including a category breakdown for SICK test data. Categories are shown with their frequencies.}
  \label{sresultstable}
\end{table} 

\begin{table*}[htp]
  \centering\small
  \begin{tabular}{l@{~~~}cl}
    \toprule
  The patient is being helped by the doctor	& \ii{entailment} & The doctor is helping the patient (\textsc{Passive})\\
    A little girl is playing the violin on a beach & \ii{contradiction} &	There is no girl playing the violin on a beach (\textsc{Neg})\\
    
    The yellow dog is drinking water from a bottle& \ii{contradiction} &	The yellow dog is drinking water from a pot  (\textsc{Subst})\\
        A woman is breaking two eggs in a bowl & \ii{neutral} &A man is mixing a few ingredients in a bowl (\textsc{MultiEd})\\
        Dough is being spread by a man & \ii{neutral} & A woman is slicing meat with a knife (\textsc{Diff})\\
    \bottomrule
  \end{tabular}
  \caption{\label{examplesofsickdata}Examples of each category used in error analysis from the SICK test data. }
\end{table*}

\paragraph{Results} Despite the small amount of high quality training data available and the lack of resources for learning lexical relationships, the results (Table~\ref{sresultstable}) show that our tree-structured models perform competitively on textual entailment, beating a strong baseline. Neither model reached the performance of the winning system (84.6\%), but the TreeRNTN did exceed that of eight out of 18 submitted systems, including several which used sophisticated hand-engineered features and lexical resources specific to the version of the entailment task at hand. 

To better understand our results, we manually annotated a fraction of the SICK test set, using mutually exclusive categories for passive/active alternation pairs (\textsc{Passive}), pairs differing only by the presence of negation (\textsc{Neg}), pairs differing by a single word or phrase substitution (\textsc{Subst}), pairs differing by multiple edits (\textsc{MultiEd}), and pairs with little or no content word overlap (\textsc{Diff}). Examples of each are in Table \ref{examplesofsickdata}. We annotated 100 random examples to judge the frequency of each category, and  continued selectively annotating until each category contained at least 25. We also use the category \textsc{Short} for pairs in which neither sentence contains more than ten words.
 
The results (Table \ref{examplesofsickdata}) show that the TreeRNTN performs especially strongly in the two categories which pick out specific syntactic configurations, \textsc{Passive} and \textsc{Neg}, suggesting that that model has learned to encode the relevant structures well. It also performs fairly on \textsc{Subst}, which most closely parallels the lexical entailment inferences addressed in \S\ref{sec:quantifiers}. In addition, none of the models perform dramatically better on the \textsc{Short} pairs than on the rest of the data, suggesting that the performance decay observed in \S\ref{sec:recursion} may not impact models trained on typical natural language text.

It is known that a model can perform well on SICK (like other natural language inference corpora) without taking advantage of compositional syntactic or semantic structure \cite{marelli2014semeval}, and our summing baseline model is powerful enough to do this. Our tree models nonetheless perform substantially better, and we remain confident that given sufficient data, it should be possible for the tree models, and not the summing model, to learn a truly high-quality solution.

\section{Discussion and conclusion}\label{sec:discussion}

This paper first evaluates two recursive models on three natural language inference 
tasks over clean artificial data, covering the 
core relational algebra of natural logic with entailment and
exclusion, recursive structure, and quantification. 
We then show that the same models can learn to
perform an entailment task on natural language. The results suggest that TreeRNTNs,
and potentially also TreeRNNs, can learn to faithfully reproduce logical inference behaviors from
reasonably-sized training sets. These positive results are
promising for the future of learned representation models in the
applied modeling of compositional semantics.

Some questions about the abilities of these models remain open. Even
the TreeRNTN falls short of perfection in the recursion experiment, with
performance falling off steadily as the size of the expressions grows. It
remains to be seen whether these deficiencies are limiting in practice,
and whether they can be overcome with
stronger models or learning techniques. In addition, interesting 
analytical questions remain about \ii{how} these models encode
the underlying logics. Neither the underlying
logical theories, nor any straightforward parameter inspection technique provides 
much insight on this point, but we hope that further experiments may reveal 
structure in the learned parameters or the representations they produce.

Our SICK experiments similarly only begin to reveal the potential of these models to learn to 
perform complex semantic inferences from corpora, and there is ample room to develop our understanding
using new and larger sources of natural language data. Nonetheless, the rapid progress the field 
has made with these models in recent years provides ample reason to be optimistic that 
learned representation models can be trained to
meet all the challenges of natural language semantics.

\subsubsection*{Acknowledgments}

We thank Jeffrey Pennington, Richard Socher, and audiences at CSLI, Nuance, and BayLearn, as well as Neha Nayak for developing the SICK collapsing technique.

\bibliographystyle{acl}
\bibliography{MLSemantics} 

\end{document}